\documentclass[letterpaper]{article}
\usepackage{aaai19}
\usepackage{times}
\usepackage{helvet}
\usepackage{courier}
\usepackage{url}
\usepackage{graphicx}
\usepackage{amssymb}
\usepackage{amsmath}
\usepackage{pgfplots}
\usepackage{tikz}
\usepackage{tabularx}
\usepackage{booktabs}
\usepackage{arydshln}
\usepackage{xcolor}
\usepackage{todonotes}
\usepackage{enumitem}
\title{
Learning Semantic Representations for Novel Words:\\
Leveraging Both Form and Context}
\author{
Timo Schick \\ Sulzer GmbH \\ Munich, Germany \\ \url{timo.schick@sulzer.de} \And Hinrich Sch\"utze 
\\ Center for Information and Language Processing \\ LMU Munich, Germany \\ \url{inquiries@cislmu.org}}
\definecolor{c-w2v}{cmyk}{1,0.3968,0,0.2588} 
\definecolor{c-both}{cmyk}{0,0.6175,0.8848,0.1490} 
\definecolor{c-alc}{cmyk}{0.1127,0.6690,0,0.4431} 
\definecolor{c-intra}{cmyk}{0.6765,0.2017,0,0.0667} 
\definecolor{c-inter}{cmyk}{0.3081,0,0.7209,0.3255} 
\definecolor{c-avg}{cmyk}{0,0.8765,0.7099,0.3647}

\usetikzlibrary{calc}

\makeatletter
\def\adl@drawiv#1#2#3{%
        \hskip.5\tabcolsep
        \xleaders#3{#2.5\@tempdimb #1{1}#2.5\@tempdimb}%
                #2\z@ plus1fil minus1fil\relax
        \hskip.5\tabcolsep}
\newcommand{\cdashlinelr}[1]{%
  \noalign{\vskip\aboverulesep
           \global\let\@dashdrawstore\adl@draw
           \global\let\adl@draw\adl@drawiv}
  \cdashline{#1}
  \noalign{\global\let\adl@draw\@dashdrawstore
           \vskip\belowrulesep}}
\makeatother

\newcommand\namecite[1]{\citeauthor{#1} (\citeyear{#1})}

\newcounter{notecounter}
\newcommand{\enotesoff}{\long\gdef\enote##1##2{}}

\enotesoff

\def\eqref#1{Eq.~\ref{eqn:#1}}

\begin{document}
\maketitle

\begin{abstract}
Word embeddings are a key component of high-performing
natural language processing (NLP) systems, but it remains a
challenge to learn \emph{good representations  for novel
words} on the fly, i.e., for words that
did not occur in the training data. The general problem
setting is that word embeddings are induced on an unlabeled training
corpus and then a model is trained that embeds novel words into
this induced embedding space.
Currently, two approaches for
learning
embeddings of novel words
exist: (i) learning an embedding from the novel word's
\emph{surface-form}
(e.g., subword $n$-grams)
and
(ii) learning an embedding from the \emph{context} in which it occurs.
In this paper, we propose an architecture that leverages
both sources of information -- surface-form and context --
and show that it results in large increases in embedding quality.
Our architecture  obtains state-of-the-art
results on the Definitional Nonce  and  Contextual
Rare Words datasets.
As input, we only require an embedding set and an
unlabeled corpus for training our architecture to produce
embeddings appropriate for the induced embedding space.
Thus, our model can easily be integrated into any existing NLP system and enhance its
capability to handle novel words.
\end{abstract}

\section{Introduction}

Distributed word representations (or embeddings) are a foundational aspect of many natural language processing systems; they have successfully been used for a wide variety of different tasks \cite{goldberg2016primer}. The idea behind  embeddings is to assign to each word a low-dimensional, real-valued vector representing its meaning. In particular, neural network based approaches such as the skipgram and cbow models introduced by \namecite{Mikolov2013} have gained increasing popularity over the last few years. 

Despite their success, an important problem with current
approaches to learning embeddings is that they require many observations of a word for its embedding to become reliable; as a consequence, they struggle with small corpora and infrequent words \cite{ataman2018compositional}. Furthermore, as models are typically trained with a fixed vocabulary, they lack the ability to assign vectors to novel, out-of-vocabulary (OOV) words once training is complete.

In recent times, several ways have been proposed to overcome
these limitations and to extend word embedding models with
the ability to obtain representations of previously unseen
words on the fly. These approaches can roughly be divided
into two directions: (i) the usage of \emph{subword
  information}, i.e., exploiting information that can be
extracted from the surface-form of the word and (ii)
the usage of \emph{context information}. The first
direction aims to obtain good embeddings for novel words by
looking at their characters \cite{pinter2017mimicking},
morphemes
\cite{lazaridou2013compositional,luong2013better,cotterell2016morphological}
or $n$-grams
\cite{wieting2016charagram,bojanowski2016enriching,ataman2018compositional,salle2018incorporating}. Naturally, this direction is especially well-suited for languages with rich morphology \cite{gerz2018language}. The
second, context-based direction tries to infer embeddings
for novel words from the words surrounding them
\cite{lazaridou2017multimodal,herbelot2017high,khodak2018carte}. 
Both directions show promising results on various
benchmarks. However, for both purely surface-form-based and purely
context-based approaches, there are many cases in which they
are highly unlikely to succeed in obtaining meaningful
embeddings. As an example, suppose that
we encounter
the following three words -- 
highlighted in bold letters -- as novel words in the given contexts:

\setlist[enumerate]{align=left, leftmargin=* } 
\begin{enumerate}[label={(\arabic*)}]
\item We should write no one off as being \textbf{unemployable}.
\item A \textbf{cardigan} is a knitted jacket or sweater with buttons up the front.
\item Unlike the grapefruit, the \textbf{pomelo} has very little importance in the marketplace.
\end{enumerate}
In sentence~(1), the context is of almost no help for
determining the meaning of the novel word, but we can deduce
its meaning without great difficulty from an analysis of the
morphemes ``un'', ``employ'' and ``able''. For sentence~(2),
the reverse is true: While the novel word's morphemes give no
indication that it is a piece of clothing, this information
can easily be derived from the context in which it
occurs. Perhaps most interesting is sentence~(3):
Both the close occurrence of the word ``grapefruit'' and the
fact that the novel word's morphemes resemble words like ``pome'', ``pomegranate''
and ``melon'' are indicative of the fact that it may be some sort of fruit. While none of those indicators may be strong enough on its own, their combination gives a pretty strong clue of the word's meaning.

As all three of the above sentences demonstrate, for an approach to cover a wide range of novel words, it is essential to make use of all available information. In this work, we therefore propose an architecture that, given a new word, captures both its subword structure and all available context information and combines them to obtain a high-quality embedding. To this end, we first infer two distinct embeddings, one incorporating the word's inner structure and one capturing its context, and then combine them into a unified word embedding. Importantly, both embeddings and their composition function are learned jointly, allowing each embedding to rely on its counterpart whenever its available information is not sufficient. In a similar fashion to work by \namecite{pinter2017mimicking} and \namecite{khodak2018carte}, our approach is not trained from scratch, but instead makes use of preexisting word embeddings and aims to reconstruct these embeddings. This allows for a much faster learning process and enables us to easily combine our approach with any existing word embedding model, regardless of its internal structure. 

Our approach is able to generate embeddings for OOV words
even from only a single observation with high accuracy in
many cases and outperforms previous work on the
Definitional Nonce dataset \cite{herbelot2017high} and the
Contextual Rare Words dataset \cite{khodak2018carte}. To the
best of our knowledge, this is the first work that jointly uses
surface-form and context information to obtain
representations for novel words.

In summary, our contributions are as follows:
\begin{itemize}
  \item We propose a new model for learning embeddings for
    novel words that leverages both surface-form and
    context.
    \item We demonstrate that this model outperforms prior
      work -- which only used one of these two sources of
      information -- by a large margin.
\item Our model is designed in a way which allows it to easily be integrated into existing
  systems.
It therefore has the potential to 
  enhance the
capability
of any NLP system that uses distributed word representations to handle novel words. 
      
\end{itemize}

\section{Related Work}

Over the last few years, many ways have been proposed to generate embeddings for novel words; we highlight here only the ones most relevant to our work.

As shown by \namecite{lazaridou2017multimodal}, one of the simplest context-based methods to obtain embeddings for OOV words is through summation over all embeddings of words occurring in their contexts. \namecite{herbelot2017high} show that with some careful tuning of its hyperparameters, the skipgram model by \namecite{Mikolov2013} can not only be used to assign vectors to frequent words, but also does a decent job for novel words; they refer to their tuned version of skipgram as \emph{Nonce2Vec}.
Very recently, \namecite{khodak2018carte} introduced the \emph{A La Carte} embedding method that, similar to the summation model by \namecite{lazaridou2017multimodal}, averages over all context words. Subsequently, a linear transformation is applied to the resulting embedding, noticeably improving results on several datasets.

In the area of subword-based approaches, \namecite{luong2013better} make use of morphological structure and use a recurrent neural network to construct word embeddings from embeddings assigned to each morpheme. Similarly, \namecite{lazaridou2013compositional} try several simple composition functions such as summation and multiplication to acquire word embeddings from morphemes. Both approaches, however, rely on external tools to obtain a segmentation of each word into morphemes.
For this reason, another direction chosen by several authors is to resort to $n$-grams instead of morphemes \cite{wieting2016charagram,ataman2018compositional}. The \emph{fastText} model introduced by \namecite{bojanowski2016enriching} is basically an extension of the skipgram model by \namecite{Mikolov2013} which, instead of directly learning vectors for words, assigns vectors to character $n$-grams and represents each word as the sum of its $n$-grams. In a similar fashion, \namecite{salle2018incorporating} incorporate $n$-grams and morphemes into the \emph{LexVec} model \cite{salle2016matrix}. A purely character-based approach was taken by \namecite{pinter2017mimicking} who, given a set of reliable word embeddings, train a character-level bidirectional LSTM \cite{hochreiter1997long} to reproduce these embeddings. 
As it learns to mimic a set of given embeddings, the authors call their model \emph{Mimick}.

\section{The Form-Context Model}

As previously demonstrated, for both purely context-based
approaches and approaches that rely entirely on surface-form
information, there are cases in which it is almost
impossible to infer a high-quality embedding for a novel
word. We now show how this issue can be overcome by
combining the two approaches into a unified model.  To this
end, let $\Sigma$ denote an alphabet and let $\mathcal{V}
\subset \Sigma^*$ be a finite set of words. We assume that
for each word in $\mathcal{V}$, we are already provided with
a corresponding word embedding. That is, there is some
function $e: \mathcal{V} \rightarrow \mathbb{R}^k$ where $k
\in \mathbb{N}$ is the dimension of the embedding space and
for each word $\mathbf{w} \in \mathcal{V}$, $e(\mathbf{w})$
is the embedding assigned to $\mathbf{w}$. This embedding
function may, for example, be obtained using the skipgram
algorithm of \namecite{Mikolov2013}.

Given the embedding function $e$, the aim of our model is to
determine high-quality embeddings for new words ${\mathbf{w}
  \in \Sigma^* \setminus \mathcal{V}}$, even if they are
observed only in a single context. Let $\textbf{w} = w_1
\ldots w_l$, $l > 0$ (i.e., $\textbf{w}$ has a length of $l$
  characters) and let $\mathcal{C} = \{C_1, \ldots, C_m\}$,
  $m > 0$ be the \emph{context set} of $\textbf{w}$, i.e., a set of contexts in which $\textbf{w}$ occurs. That is, for all $i \in \{1, \ldots, m\}$, 
\[
C_i = \{ \textbf{w}_i^1, \ldots, \textbf{w}_i^{k_i}\}
\]
is a multiset of words over $\Sigma$ with $k_i \in \mathbb{N}$ and there is some $j \in \{1,
\ldots, k_i\}$ such that $\textbf{w}_i^j = \textbf{w}$. We
compute two distinct embeddings, one using only the surface-form
information of $\textbf{w}$ and one using only the
context set $\mathcal{C}$, and then combine both embeddings to
obtain our final word representation.

We first define the \emph{surface-form embedding} that is
obtained making use only of the word's letters $w_1, \ldots,
w_l$ and ignoring the context set $\mathcal{C}$.
To this end, we pad the word with special start and end tokens $w_0 = \langle s\rangle$, $w_{l+1} = \langle e \rangle$ and define the multiset
\[
S_\textbf{w} = \bigcup_{n=n_\text{min}}^{n_\text{max}} \bigcup_{i = 0}^{l+2-n} \{ w_i  w_{i+1}  \ldots  w_{i+n-1} \}
\]
consisting of all $n$-grams contained within $\mathbf{w}$ for which $n_\text{min} \leq n \leq n_\text{max}$. For example, given $n_\text{min} = 2, n_\text{max} = 3$, the $n$-gram set for the word \emph{pomelo} is
\begin{align*}
S_\text{pomelo} =\ & \{ \langle s \rangle\text{p}, \text{po}, \text{om}, \text{me}, \text{el}, \text{lo}, \text{o}\langle e \rangle \} \\ & \cup \{ \langle s \rangle \text{po}, \text{pom}, \text{ome}, \text{mel}, \text{elo}, \text{lo} \langle e \rangle  \}.
\end{align*}

To transform the $n$-grams into our semantic space, we
introduce an \emph{$n$-gram embedding function}
$e_\text{ngram}: \Sigma^* \rightarrow \mathbb{R}^k$ which
assigns an embedding to each $n$-gram. In a fashion similar
to \namecite{bojanowski2016enriching}, we then define the
surface-form embedding of $\textbf{w}$ to be the average of all its $n$-gram embeddings:
\[
v_{(\textbf{w},\mathcal{C})}^\text{form} = \frac{1}{|S_\textbf{w}|} \sum_{s \in S_\textbf{w}} e_\text{ngram}(s).
\]
Unlike the word-based embedding function $e$, we do not assume $e_\text{ngram}$ to be given, but instead treat it as a learnable parameter of our model, implemented as a lookup table.

Complementary to this first embedding based solely on
surface-form information, we also define a \emph{context
  embedding}. This embedding is constructed only from the
context set $\mathcal{C}$ in which $\mathbf{w}$ is observed,
making no use of its characters. Analogous to the surface-form
embedding, we obtain this embedding by averaging over all
context words:

\[
v_{(\textbf{w}, \mathcal{C})}^\text{context} = \frac{1}{c}
\  \sum_{C \in \mathcal{C}} \ \sum_{\mathbf{w}' \in C \cap \mathcal{V}} e(\mathbf{w}')
\]
where $c = \sum_{C \in \mathcal{C}} |C \cap \mathcal{V}|$ is the total number of words in $\mathcal{C}$ for which embeddings exist. In accordance with results reported by \namecite{khodak2018carte}, we found it helpful to apply a linear transformation to the so-obtained embedding, resulting in the final context embedding
\[
\hat{v}_{(\textbf{w}, \mathcal{C})}^\text{context} = A \cdot {v}_{(\textbf{w}, \mathcal{C})}^\text{context}
\]
with $A \in \mathbb{R}^{k \times k}$ being a learnable parameter of our model.

We finally combine both embeddings to obtain a joint embedding $v_{(\mathbf{w}, \mathcal{C})}$ for $\textbf{w}$. The perhaps most intuitive way of doing so is to construct a linear combination
\[
v_{(\textbf{w}, \mathcal{C})} = \alpha \cdot \hat{v}_{(\textbf{w}, \mathcal{C})}^\text{context} + (1-\alpha) \cdot v_{(\textbf{w}, \mathcal{C})}^\text{form}.
\]
In one configuration of our model,
$\alpha \in [0,1]$
is a single learnable parameter. We call this version the \emph{single-parameter model}.

However, it is highly unlikely that there is a single value of $\alpha$ that works well for every pair $(\mathbf{w}, \mathcal{C})$ -- after all, we want $\alpha$ to be large whenever $\mathcal{C}$ helps in determining the meaning of $\mathbf{w}$ and, conversely, want it to be small whenever $S_\mathbf{w}$ is more helpful. We therefore also consider a second, more complex architecture in which the value of $\alpha$ directly depends on the two embedding candidates. This is achieved by setting 
\[
\alpha = \sigma(w^\top [{v}_{(\textbf{w}, \mathcal{C})}^\text{context}\circ{v}_{(\textbf{w}, \mathcal{C})}^\text{form}] + b)
\]
with $w\in \mathbb{R}^{2k}$, $b \in \mathbb{R}$ being
learnable parameters of our model, $\circ$ denoting vector
concatenation and $\sigma$ denoting the sigmoid function.
We call this version of the model the \emph{gated model}
since we can view $\alpha$ as a gate in this case.

In addition to the single-parameter and gated models, we also
tried several more sophisticated composition functions,
including a variant where $\alpha$ is computed using a
multi-layer neural network and another variant with $\alpha
\in [0,1]^k$ being a component-wise weighing
parameter. Furthermore, we experimented with an iterative
procedure that refines the combined embedding over multiple
iterations by adjusting the composition based on embeddings
obtained from previous iterations. In our
experiments, however, none of these modifications did
consistently improve the model's performance, so we do not
investigate them in detail here.

As
it combines context and surface-form embeddings, we refer
to the final embedding $v_{(\mathbf{w}, \mathcal{C})}$
obtained using the composition function
(in both single-parameter and gated models)
as a
\emph{form-context word embedding}. The overall architecture
of our model is shown schematically in
Figure~\ref{model-architecture}.

\tikzset{
  block-dashed/.style={rectangle, inner sep=0.1cm, minimum height=6ex, text centered, draw=black, dashed},
  block/.style={rectangle, inner sep=0.1cm, minimum height=6ex, text centered},
  circ/.style={circle, text centered, minimum height=6ex, draw=black,
      text height=1.5ex,
      text depth=.25ex},
  arrow/.style={draw,->,>=stealth},
}
\begin{figure}
\centering
\begin{tikzpicture}[]
    \node [block] (wdots) {$\ldots$};
    \node [block-dashed, left=0.1cm of wdots] (w1) {$e_\text{ngram}(s_1)$};
    \node [block-dashed, right=0.1cm of wdots] (wn) {$e_\text{ngram}(s_{|S_\textbf{w}|})$};
    
    \node [block, right=0.5cm of wn] (w-c) {$e({\mathbf{w}_{1}^1})$};
    \node [block, right=0.1cm of w-c] (w-cdots) {$\ldots$};
    \node [block, right=0.1cm of w-cdots] (w-1) {$e(\mathbf{w}_{m}^{k_m})$};

    \node [circ, below=0.5cm of w-cdots] (sig2) {$\text{avg}$};
    
    \node [circ, dashed, below=0.4cm of sig2] (atrans) {$A$}; 
    \node [circ, below=2cm of wdots] (sig1) {$\text{avg}$};
       
    \node [circ, yshift=-1.2cm, dashed] (compose) at ($(sig1)!0.5!(sig2)$) {$\alpha$};
    
    \node [block, below=1cm of compose] (vw) {${v}_{(\textbf{w}, \mathcal{C})}$};
    
    \path [arrow] (w1) -- (sig1);
    \path [arrow] (wdots) -- (sig1);
    \path [arrow] (wn) -- (sig1);
    
    \path [arrow] (sig2) -- (atrans);

    \path [arrow] (w-c) -- (sig2);
    \path [arrow] (w-cdots) -- (sig2);
    \path [arrow] (w-1) -- (sig2);  
    
    \path [arrow] (sig1) -- (compose); 
    \path [arrow] (atrans) -- (compose); 
    
    \path [arrow] (compose) -- (vw);      
\end{tikzpicture}
\caption{Schematic representation of the form-context word embedding architecture. Learnable parameters of the model are indicated by dashed lines.}
\label{model-architecture}
\end{figure}
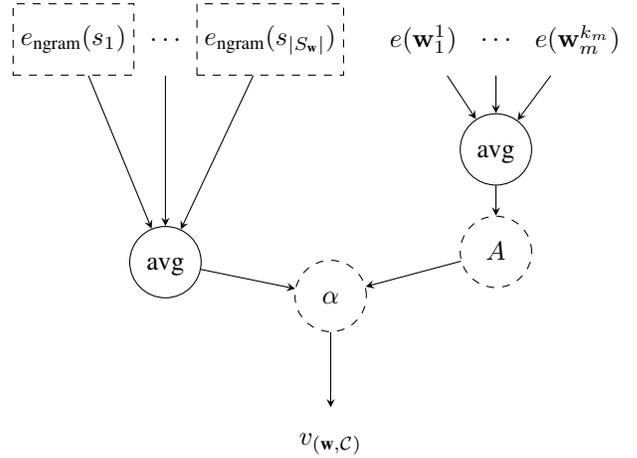

For training of our model and estimation of its learnable parameters, we require the embedding function $e$ and a training corpus $\mathcal{T}$, consisting of pairs $(\textbf{w}, \mathcal{C})$ as above. Given a batch $\mathcal{B} \subset \mathcal{T}$ of such training instances, we then aim to minimize the function
\[
L_\mathcal{B} = \frac{1}{|\mathcal{B}|}\sum_{(\textbf{w}, \mathcal{C}) \in \mathcal{B}} \|v_{(\textbf{w},\mathcal{C})} - e(\textbf{w})\|^2
\]
i.e., our loss function is the squared error between the embedding assigned to $\textbf{w}$ by $e$ and the embedding constructed by our model.

\section{Experimental Setup}

\subsection{Datasets}

We evaluate our model on two different datasets: the Definitional Nonce (DN) dataset introduced by \namecite{herbelot2017high} and the Contextual Rare Words (CRW) dataset of \namecite{khodak2018carte}. The DN dataset consists of $300$ test and $700$ train words; for each word, a corresponding definitional sentence extracted from Wikipedia is provided. The authors also provide $400$-dimensional embedding vectors for a set of 259,376 words, including the test and train words. These embeddings were obtained using the skipgram algorithm of \namecite{Mikolov2013}. On the DN dataset, our model can be evaluated by training it with all given word vectors -- except for the test set -- and then comparing the inferred embeddings for the test words with their actual embeddings. 

Our second benchmark, the CRW dataset, is based on the Rare Words dataset by \namecite{luong2013better} and contains $562$ pairs of rare words along with human similarity judgments. For each rare word, $255$ corresponding sentences are provided. In contrast to the sentences of the DN dataset, however, they are sampled randomly from the Westbury Wikipedia Corpus (WWC) \cite{shaoul2010westbury} and, accordingly, do not have a definitional character in many cases. \namecite{khodak2018carte} also provide a set of $300$-dimensional word embeddings which, again, can be used to train our model. We may then compare the similarities of the so-obtained embeddings with the given similarity scores.
As the CRW dataset comes without development data on which hyperparameters might be optimized, we extend the dataset by creating our own development set.\footnote{Our development set is publicly available at \url{https://github.com/timoschick/form-context-model}} To this end, we sample 550 random pairs of words from the Rare Words dataset, with the only restrictions that (i) the corresponding rare words must not occur in any of the pairs of the CRW dataset and (ii) they occur in at least $128$ sentences of the WWC. We then use the WWC to obtain randomly sampled contexts for each rare word in these pairs.

\subsection{Model Setup and Training}

For our evaluation on both datasets, we use the WWC to obtain the contexts required for training; the same corpus was also used by \namecite{herbelot2017high} and \namecite{khodak2018carte} for training of their models.

To construct our set of training instances, we restrict
ourselves to words occurring at least $100$ times in the
WWC. We do so because embeddings of words occurring too
infrequently generally tend to be of rather low quality. We therefore have no clear evaluation in these cases as our
model 
may do a good job at constructing an embedding for an infrequent word, but it may
be far from the word's original, low-quality embedding.
Let $\textbf{w} \in \mathcal{V}$ be a word and let
$c(\textbf{w})$ denote the number of occurrences of
$\textbf{w}$ in our corpus.
For each iteration over our dataset, we create $n(\textbf{w})$ training instances $\{(\textbf{w}, \mathcal{C}_1), \ldots, (\textbf{w}, \mathcal{C}_{n(\textbf{w})}) \}$ from this word, where
\[
n(\textbf{w}) = \min(\lfloor\frac{c(\textbf{w})}{100}\rfloor,5).
\] 
The number $n(\textbf{w})$ is designed to put
a bit more emphasis on very frequent words as we assume
that, up to a certain point, the quality of a word's
embedding increases with its frequency. For each $i \in \{1, \ldots, n(\textbf{w}) \}$, the context set $\mathcal{C}_i$ is constructed by sampling 20 random sentences from our corpus that contain $\textbf{w}$.

For surface-form embeddings, we set $n_\text{min} = 3$ and $n_\text{max} = 5$. We only consider $n$-grams that occur in at least $3$ different words of our training corpus; every other $n$-gram is replaced by a special $\langle \textit{unk}\rangle$ token. We initialize all parameters as described by \namecite{glorot2010understanding} and use a batch size of $64$ examples per training step. Training is performed using the Adam optimizer \cite{kingma2014adam} and a learning rate of $0.01$. For training of our model with the embeddings provided by \namecite{herbelot2017high}, both the learning rate and the number of training epochs is determined using the train part of the DN dataset, searching in the range $\{0.1, 0.01, 0.001\}$ and $\{1, \ldots, 10 \}$, respectively. As we assume both the quality and the dimension of the original embeddings to have a huge influence on the optimal parameters for our model, we separately optimize these parameters for training on the embeddings by \namecite{khodak2018carte} using our newly constructed development set. In all of the experiments described below, we use the cosine distance to measure the similarity between two embedding vectors.

\section{Evaluation}

To evaluate the quality of the representations obtained using our method, we train our model using the embeddings of \namecite{herbelot2017high} and compare the inferred embeddings for all words in the DN test set with their actual embeddings. For this comparison, we define the rank of a word $\textbf{w}$ to be the position of its actual embedding $e(\textbf{w})$ in the list of nearest neighbors of our inferred embedding $v_{(\mathbf{w},\mathcal{C})}$, sorted by similarity in descending order. That is, we simply count the number of words whose representations are more similar to the embedding assigned to $\textbf{w}$ by our model than its original representation. For our evaluation, we compute both the median rank and the mean reciprocal rank (MRR) over the entire test set.

\newcolumntype{R}{>{\raggedleft\arraybackslash}X}
\begin{table}
  \centering
  {
\begin{tabularx}{\linewidth}{Xcrr}
\toprule
\textbf{Model} & \textbf{Type} & \multicolumn{1}{l}{\textbf{Median Rank}} & \multicolumn{1}{l}{\textbf{MRR}} \\
\midrule
Mimick  & S & 85573\phantom{.5} & 0.00006 \\
Skipgram & C & 111012\phantom{.5} & 0.00007 \\
Additive & C & 3381\phantom{.5} & 0.00945 \\
Nonce2Vec & C & 623\phantom{.5} & 0.04907 \\
A La Carte & C & 165.5 & 0.07058 \\
\midrule
surface-form & S & 404.5 & 0.12982 \\
context & C & 184\phantom{.5} & 0.06560 \\
single-parameter & S\&C & 55\phantom{.5} & 0.16200 \\
gated & S\&C& \textbf{49}\phantom{.5} & \textbf{0.17537} \\
\bottomrule
\end{tabularx}}
\caption{Results of various approaches on the DN
  dataset. The ``Type'' column indicates whether the model
  makes use of surface-form information (S) or context
  information (C).
Results are shown for \emph{single-parameter} and \emph{gated}
configurations of the form-context model.
\label{results-nd}}
\end{table}

The results of our model and various other approaches are
shown in Table~\ref{results-nd}. Scores for the original
skipgram algorithm, the Nonce2Vec model and an additive
baseline model that simply sums over all context embeddings
are adopted from \namecite{herbelot2017high}, the result of
the A La Carte embedding method is the one reported by
\namecite{khodak2018carte}. To obtain results for the Mimick
model, we used the original implementation by
\namecite{pinter2017mimicking}. Recall that
we distinguish between the single-parameter model, in which
the composition coefficient $\alpha$ is a single
learnable parameter, and the gated model, in
which $\alpha$ depends on the two embeddings.
To see whether any potential
improvements over previous approaches are indeed due to our
combination of surface-form and context information and
not just due to differences in the models themselves, we also
report scores obtained using only the surface-form and only
the context
parts of our model, respectively.

As can be seen, using only surface-form information results
in a comparatively high MRR, but the obtained median rank is
rather bad. This is due to the fact that the surface-form
model assigns very good embeddings to words whose meaning
can be inferred from a morphological analysis, but
completely fails to do so for most other words. The context
model, in contrast, works reasonably well for almost all
words but only infrequently achieves single-digit ranks. The
combined form-context model clearly outperforms not only the
individual models, but also beats all previous
approaches. Interestingly, this is even the case for the
single-parameter model,
in which $\alpha$ is constant across all words. The optimal value of $\alpha$ learned by this model is $0.19$, showing a clear preference towards surface-form embeddings.

The gated configuration further
improves the model's performance noticeably. Especially the
median rank of $49$ achieved using the gated model
architecture is quite remarkable: Considering that the
vocabulary consists of 259,376 words, this means that for
50\% of the test set words, at most 0.019\% of all words
in the vocabulary are more similar to the inferred embedding
than the actual embedding. Similar to the single-parameter model, the average value of $\alpha$ over the entire test set for the gated model is $0.20$, with individual values ranging from $0.07$ to $0.41$. While this shows how the gated model learns to assign different weights based on word form and context, the fact that it never assigns values above $\alpha = 0.41$ -- i.e., it always relies on the surface-form embedding to a substantial extent -- indicates that the model may even further be improved through a more elaborate composition function.

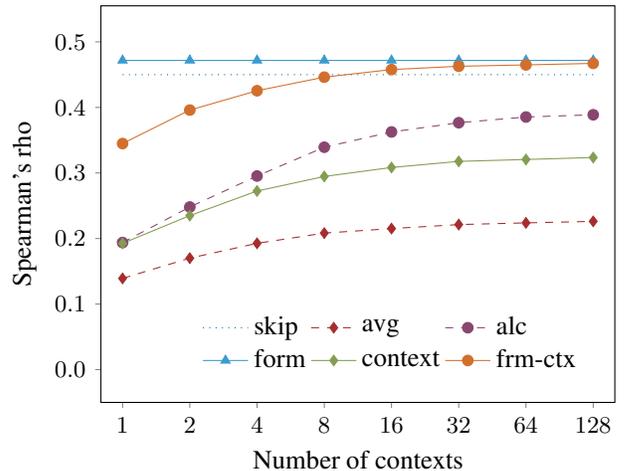
\begin{figure}
\centering
\begin{tikzpicture}
\begin{axis}[
	cycle list name=color list,
	xlabel={Number of contexts},
		ylabel={Spearman's rho},
    ymin = -0.05,
    ymax = 0.555,
    xmin = 0.8,
    xmax = 160,
    xmode = log,
    xtick pos=left,
    ytick pos=left,
    log basis x={2},
        ylabel near ticks,
        xlabel near ticks,
            ytick={0, 0.1, 0.2, 0.3, 0.4, 0.5},
            yticklabels={$0.0$, $0.1$, $0.2$, $0.3$, $0.4$, $0.5$},
    xtick = data,
    tick align=outside,
          major tick length=0.075cm,
    width = \linewidth,
    height = 0.3\textheight,
    log ticks with fixed point,
    x tick label style={/pgf/number format/1000 sep=\,},
    legend style={at={(0.95,0.05)},anchor=south east, draw=none},
    legend cell align=left,
    legend columns=3,
    tick label style={font=\footnotesize}
]

\addplot[mark=none, dotted, color=c-w2v] coordinates {
(1,0.45)
(2,0.45)
(4,0.45)
(8,0.45)
(16,0.45)
(32,0.45)
(64,0.45)
(128,0.45) 
};
\addlegendentry{skip};

\addplot+[mark=diamond*, dashed, mark options={solid}, color=c-avg] coordinates {
(1,0.1392)
(2,0.1701)
(4,0.1927)
(8,0.2083)
(16,0.2152)
(32,0.2213)
(64,0.2239)
(128,0.2262)
};
\addlegendentry{avg};

\addplot+[mark=*, dashed, mark options={solid}, color=c-alc] coordinates {
(1,0.1939)
(2,0.2481)
(4,0.2954)
(8,0.3393)
(16,0.3626)
(32,0.3767)
(64,0.3854)
(128,0.3888)
};
\addlegendentry{alc};

\addplot+[mark=triangle*, color=c-intra] coordinates {
(1,0.4718)
(2,0.4718)
(4,0.4718)
(8,0.4718)
(16,0.4718)
(32,0.4718)
(64,0.4718)
(128,0.4718)
};
\addlegendentry{form};

\addplot+[mark=diamond*, color=c-inter] coordinates {
(1,0.1926)
(2,0.2350)
(4,0.2725)
(8,0.2947)
(16,0.3084)
(32,0.3177)
(64,0.3206)
(128,0.3236)
};
\addlegendentry{context};

\addplot+[mark=*, color=c-both] coordinates {
(1,0.3449)
(2,0.3960)
(4,0.4255)
(8,0.4462)
(16,0.4578)
(32,0.4628)
(64,0.4649)
(128,0.4672)
};
\addlegendentry{frm-ctx};

\end{axis}
\end{tikzpicture}
\caption{Results on the CRW dataset by
  \cite{khodak2018carte} for the averaging baseline (avg), A
  La Carte (alc), the surface-form model (form), the context
  model (context) and the combined form-context model in its gated version (frm-ctx) as well as for the skipgram algorithm (skip) when trained on all 255 contexts}
\label{results-crw}
\end{figure}

As a second evaluation, we turn to the CRW dataset for which
results are shown in
Figure~\ref{results-crw}.\footnote{Results reported in
  Figure~\ref{results-crw} differ slightly from the ones by
  \namecite{khodak2018carte} because for each word pair
  $(\textbf{w}_1, \textbf{w}_2)$ of the CRW corpus, the
  authors only estimate an embedding for $\textbf{w}_2$ and
  take $e(\textbf{w}_1)$ as the embedding for
  $\textbf{w}_1$; if $\textbf{w}_1$ is not in the domain of
  $e$, a zero vector is  taken instead. In contrast, we
  simply infer an embedding for $\textbf{w}_1$ analogically
  to $\textbf{w}_2$ in the latter case.} We use Spearman's
rho as a measure of agreement between the human similarity
scores and the ones assigned by the model. As the CRW
dataset provides multiple contexts per word, we can also
analyze how modifying the number of available contexts
influences the model's performance. As can be seen, our
model again beats the averaging baseline and A La Carte by a
large margin, regardless of the number of available
contexts. Interestingly, with as little as $8$ contexts,
our model is almost on par with the original skipgram embeddings --
which were obtained using all $255$ contexts -- and even
improves upon them given $16$ or more contexts. However, it
can also be seen that the surface-form model actually
outperforms the combined model. While this may at first seem
surprising, it can be explained by looking at how the CRW
dataset was constructed: Firstly, \namecite{luong2013better}
focused explicitly on morphologically complex words when
creating the original Rare Words dataset, so the CRW dataset
contains many words such as ``friendships'', ``unannounced''
or ``satisfactory'' that are particularly well-suited for
an exclusively surface-form-based model. Secondly, the provided contexts for each word are sampled randomly, meaning that they are of much lower definitional quality than the single sentences provided in the DN dataset. Despite this bias of the dataset towards surface-form-based models, given $32$ or more contexts, the combined model performs comparable to the surface-form embeddings. However, the results clearly indicate that our model may even further be improved upon by incorporating the number and quality of the available contexts into its composition function.

Of course, we can also compare our approach to the purely
surface-form-based fastText method of
\namecite{bojanowski2016enriching}, which, however, makes no
use of the original embeddings by \namecite{khodak2018carte}. We therefore train $300$-dimensional fastText embeddings from scratch on the WWC, using the same values of $n_\text{min}$ and $n_\text{max}$ as for our model. While
the so-trained model achieves a value of $\rho = 0.496$ --
as compared to $\rho = 0.471$ for our surface-form model --
a direct comparison to our method is not appropriate as our
model's performance is highly dependent on the embeddings it
was trained from. We can, however, train our method on the
embeddings provided by fastText to allow for a fair
comparison. Doing so results in a score of $\rho = 0.508$
for the gated model when using $128$ contexts, showing that
even for word embedding algorithms that already make use of
surface-form information, our method is helpful in obtaining
high-quality embeddings for novel words. Noticeably, when
trained on fastText embeddings, the form-context  model
even outperforms the surface-form model ($\rho = 0.501$).

We also evaluate the form-context model on seven supervised sentence-level classification tasks using the \mbox{SentEval} toolkit \cite{conneau2018senteval}.\footnote{We use the MRPC, MR, CR, SUBJ, MPQA, SST2 and SST5 tasks for this evaluation.} To do so, we train a simple bag-of-words model using the skipgram embeddings provided by \namecite{khodak2018carte} and obtain embeddings for OOV words from either the form-context model, the A~La~Carte embedding method or the averaging baseline, using as contexts all occurrences of these words in the WWC. While the form-context model  outperforms all other models, it does so by only a small margin with an average accurracy of $75.34$ across all tasks, compared to accuracies of $74.98$, $74.90$ and $75.27$ for skipgram without OOV words, A~La~Carte and the averaging baseline, respecitvely. Presumably, this is because novel and rare words have only a small impact on performance in these sentence-level classification tasks.

\section{Analysis}

For a qualitative analysis of our approach, we use the gated model
trained with the embeddings provided by
\namecite{herbelot2017high}, look at the nearest neighbors
of some embeddings that it infers and investigate the
factors that contribute most to these embeddings.
We attempt to measure
the contribution of a single $n$-gram or context word to the
embedding of a word $\mathbf{w}$ by simply
computing the
cosine distance between the inferred embedding $v_{(\mathbf{w},
  \mathcal{C})}$ and the embedding obtained when removing
this specific $n$-gram or word.

For a quantitative analysis of our approach, we measure the influence of combining both models on the embedding quality of each word over the entire DN test set.

\def\tabularxcolumn#1{m{#1}}
\newcolumntype{L}{>{\raggedright\arraybackslash}X}
\begin{table}
\centering
{
\begin{tabularx}{\linewidth}{m{1cm}LLL} 
    \toprule
 & \textbf{spies} & \textbf{hygiene} & \textbf{perception} \\ 
    \midrule
\textbf{form} & pies, cakes, spied, sandwiches & hygienic, hygiene, cleansers, {hypoallergenic} & interception, interceptions, fumble, touchdowns \\ 
\cdashlinelr{1-4}
\textbf{rank} & \multicolumn{1}{r}{668} & \multicolumn{1}{r}{2} & \multicolumn{1}{r}{115}\\
\midrule
\textbf{context} & espionage, clandestine, covert, spying & hygieia, goddess, eileithyia, asklepios & sensory, perceptual, auditory, contextual \\ 
\cdashlinelr{1-4}
\textbf{rank} & \multicolumn{1}{r}{8} & \multicolumn{1}{r}{465} & \multicolumn{1}{r}{51} \\
\midrule
\textbf{frm-ctx} & espionage, spying, clandestine, covert & hygienic, hygieia, health, hygiene &  sensory, perceptual, perception, auditory \\ 
\cdashlinelr{1-4}
\textbf{rank} & \multicolumn{1}{r}{6} & \multicolumn{1}{r}{4} & \multicolumn{1}{r}{3} \\
\bottomrule
\end{tabularx}}
\caption{Nearest neighbors and ranks of selected words when
  using surface-form embeddings, context embeddings and
  gated form-context  (frm-ctx) embeddings}
\label{qualitative-analysis}
\end{table}

\subsection{Qualitative analysis}
Table~\ref{qualitative-analysis} lists the nearest
neighbors of the inferred embeddings for selected words
from the DN dataset where the context set $\mathcal{C}$
simply consists of the single definitional sentence provided.
For each embedding $v_{(\mathbf{w}, \mathcal{C})}$, Table~\ref{qualitative-analysis} also shows the rank of the actual word $\mathbf{w}$, i.e., the position of the actual embedding $e(\mathbf{w})$ in the sorted list of nearest neighbors. It can be seen that the combined model is able to find high-quality embeddings even if one of the simpler models fails to do so. For example, consider the word ``spies'' for which the surface-form model fails to find a good embedding. The reason for this becomes obvious when analyzing the contribution of each $n$-gram for the final embedding. This contribution is shown at the top of Figure~\ref{relevant-words}, where a darker background corresponds to higher contribution. It can be seen there that the high contribution of $n$-grams also occurring in the word ``pies'' -- which, while having a similar surface-form, is semantically completely different from ``spies'' --, is the primary reason for the low quality embedding. Despite this, the embeddings found by both the context model and the combined model are very close to its actual embedding.

In a similar fashion, the context model is not able to come up with a good embedding for the word ``hygiene'' from the provided definitional sentence. This sentence can be seen at the bottom of Figure~\ref{relevant-words} where, as before, words are highlighted according to their importance. While the linear transformation applied to the context embeddings helps to filter out stop words such as ``which'', ``of'' and ``the'' which do not contribute to the word's meaning, the sentence is still too complex for our model to focus on the right words. This results in the context embedding being closer to words from Greek mythology than to words related to hygiene. Again, the combined model is able to alleviate the negative effect of the context model, although it performs slightly worse than the purely surface-form-based model. For the last example provided, ``perception'', neither of the two simpler models performs particularly well: The surface-form model is only able to capture the word's part of speech whereas the context model finds semantically related words with different parts of speech. Interestingly, the form-context model is still able to infer a high-quality embedding for the word, combining the advantages of both models it is composed of. 

\begin{figure}
\centering
\begin{tikzpicture}
	\node[text width=\linewidth](spies-ngrams) {
		\colorbox{black!4}{$\langle s \rangle$sp\vphantom{yl}}
		\colorbox{black!5}{$\langle s \rangle$spi\vphantom{yl}}
		\colorbox{black!25}{$\langle s \rangle$spie\vphantom{yl}}
		\colorbox{black!12}{spi\vphantom{yl}}
		\colorbox{black!17}{spie\vphantom{yl}}
		\colorbox{black!0}{spies\vphantom{yl}}
		\colorbox{black!6}{pie\vphantom{yl}}
		\colorbox{black!22}{pies\vphantom{yl}}
		\colorbox{black!19}{pies$\langle e \rangle$\vphantom{yl}}
		\colorbox{black!3}{ies\vphantom{yl}}
		\colorbox{black!5}{ies$\langle e \rangle$\vphantom{yl}} 
		\colorbox{black!1}{es$\langle e \rangle$\vphantom{yl}}
	};
	\node[text width=\linewidth, below=0.2cm of spies-ngrams](hygiene) {
		\colorbox{black!2}{which\vphantom{yl}}
		\colorbox{black!10}{comes\vphantom{yl}}
		\colorbox{black!3}{from\vphantom{yl}}
		\colorbox{black!1}{the\vphantom{yl}}
		\colorbox{black!10}{name\vphantom{yl}}
		\colorbox{black!1}{of\vphantom{yl}}
		\colorbox{black!1}{the\vphantom{yl}}
		\colorbox{black!21}{greek\vphantom{yl}}
		\colorbox{black!25}{goddess\vphantom{yl}}
		\colorbox{black!1}{of\vphantom{yl}}
		\colorbox{black!16}{health\vphantom{yl}}
		\colorbox{black!14}{hygieia\vphantom{yl}}
		\colorbox{black!3}{is\vphantom{yl}}
		\colorbox{black!3}{a\vphantom{yl}}
		\colorbox{black!8}{set\vphantom{yl}}
		\colorbox{black!1}{of\vphantom{yl}}
		\colorbox{black!18}{practices\vphantom{yl}}
		\colorbox{black!13}{performed\vphantom{yl}}
		\colorbox{black!2}{for\vphantom{yl}}
		\colorbox{black!1}{the\vphantom{yl}}
		\colorbox{black!20}{preservation\vphantom{yl}}
		\colorbox{black!1}{of\vphantom{yl}}
		\colorbox{black!16}{health\vphantom{yl}}
	};	
\end{tikzpicture}
\caption{Importance of $n$-grams for the surface-form embedding of ``spies'' (top) and of context words for the context embedding of ``hygiene'' (bottom)
\label{relevant-words}}
\end{figure}

The values of $\alpha$ assigned to all three of the above words by the gated model show that, to some extent, it is able to distinguish between cases in which context is helpful and cases where it is better to rely on surface-form information: While the embedding for ``hygiene'' is composed with a value of $\alpha = 0.22$, both the embeddings of ``spies'' and ``perception'' put more focus on the context ($\alpha = 0.32$ and $\alpha = 0.33$, respectively). To further analyze the weights learned by our model, Table~\ref{results-word-selection} lists some exemplary words with both comparably high and low values of $\alpha$. The words with the lowest values almost exclusively refer to localities that can easily be identified by their suffixes (e.g. ``ham'', ``bury''). Among the words with high values of $\alpha$, there are many abbreviations and words that can not easily be reduced to known lemmata.

\subsection{Quantitative analysis}
\begin{figure}[t!]
\centering
\begin{tikzpicture}
\begin{axis}[
	cycle list name=color list,
	xlabel={Rank difference bucket},
		ylabel={Number of words},
    ymin = 0,
    ymax = 105,
    xmin = -4.5,
    xmax = 4.5,
    domain = -5:5,
    tick pos=left,
    log basis x={10},
        ylabel near ticks,
        xlabel near ticks,
    xtick = {-4, -3, -2, -1, 0, 1, 2, 3, 4},
    xticklabels = {-$10^4$, -$10^3$, -$10^2$, -$10\vphantom{^0}$, $0\vphantom{^0}$, $10\vphantom{^0}$, $10^2$, $10^3$, $10^4$},
    tick align=outside,
          major tick length=0.075cm,
    width = \linewidth,
    height = 0.3\textheight,
    legend style={at={(0.995,0.95)},anchor=north east, draw=none},
    legend cell align=left,
    tick label style={font=\footnotesize}
]

\addplot[mark=triangle, dotted, mark options={solid}, color=c-w2v] coordinates {
(-4,96)
(-3,31)
(-2,29)
(-1,17)
(0,63)
(1,19)
(2,23)
(3,14)
(4,7)
};
\addlegendentry{single-parameter};

\addplot+[mark=o, dashed, mark options={solid}, color=c-avg] coordinates {
(-4,98)
(-3,30)
(-2,29)
(-1,21)
(0,60)
(1,18)
(2,23)
(3,13)
(4,7)
};
\addlegendentry{gated};

\addplot +[mark=none] coordinates {(0, 0) (0, 115)};

\end{axis}

\end{tikzpicture}

\begin{tikzpicture}

\begin{axis}[
	cycle list name=color list,
	xlabel={Rank difference bucket},
		ylabel={Number of words},
    ymin = 0,
        clip = false,
    ymax = 56,
    xmin = -4.5,
    xmax = 4.5,
    domain = -5:5,
    tick pos=left,
    log basis x={10},
        ylabel near ticks,
        xlabel near ticks,
    xtick = {-4, -3, -2, -1, 0, 1, 2, 3, 4},
    xticklabels = {-$10^4$, -$10^3$, -$10^2$, -$10\vphantom{^0}$, $0\vphantom{^0}$, $10\vphantom{^0}$, $10^2$, $10^3$, $10^4$},
    tick align=outside,
          major tick length=0.075cm,
    width = \linewidth,
    height = 0.3\textheight,
    legend style={at={(0.995,0.95)},anchor=north east, draw=none},
    legend cell align=left,
    tick label style={font=\footnotesize}
]
\node[] at(axis cs:0,57) {};
\addplot[mark=triangle, dotted, mark options={solid}, color=c-w2v] coordinates {
(-4,26)
(-3,45)
(-2,53)
(-1,44)
(0,45)
(1,29)
(2,28)
(3,26)
(4,3)
}; 
\addlegendentry{single parameter};

\addplot+[mark=o, dashed, mark options={solid}, color=c-avg] coordinates {
(-4,26)
(-3,49)
(-2,54)
(-1,50)
(0,44)
(1,27)
(2,27)
(3,17)
(4,5)
};
\addlegendentry{gated};

\addplot +[mark=none] coordinates {(0, 0) (0, 56)};

\end{axis}
\end{tikzpicture}

\caption{Effect of adding the context submodel (top) and the
surface-form submodel (bottom).  The rank difference buckets
were created by applying the $d_\text{form}$ difference
function (top) and
$d_\text{context}$ difference function (bottom) to the entire DN test set.}
\label{ranks-inter}
\label{ranks-intra}
\end{figure}
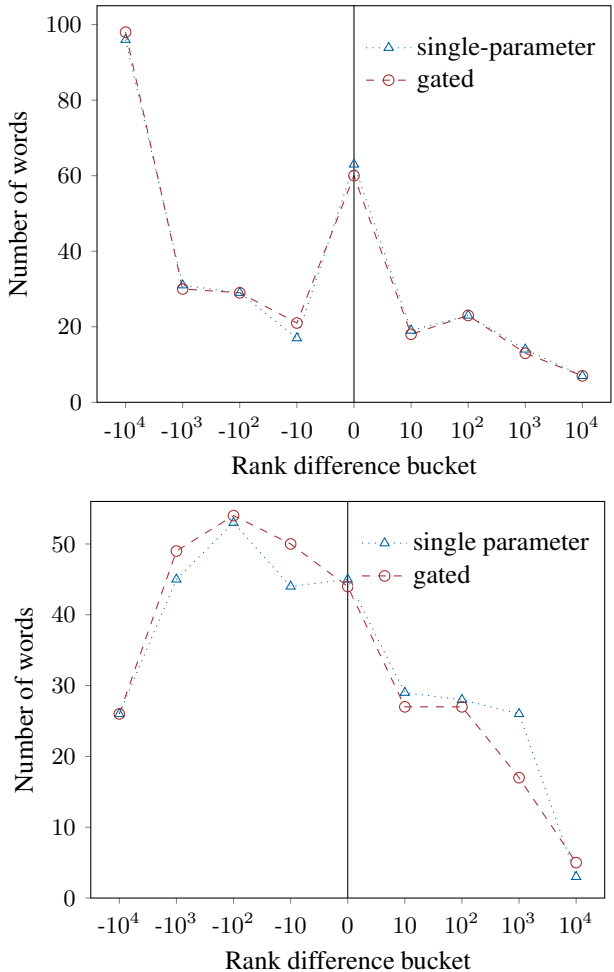

While the selected words in Table~\ref{qualitative-analysis} demonstrate cases in which the representation's quality does either improve or at least not substantially deteriorate through the combination of both embeddings, we also quantitatively analyze the effects of combining them to gain further insight into our model. To this end, let $r_\text{form}(\mathbf{w})$, $r_\text{context}(\mathbf{w})$ and $r_\text{frm-ctx}(\mathbf{w})$ denote the rank of a word $\mathbf{w}$ when the surface-form model, the context model and the form-context model is used, respectively. We measure the influence of combining both models by computing the differences
\[
d_m(\mathbf{w}) =  r_\text{frm-ctx}(\mathbf{w}) - r_m(\mathbf{w})
\]
for each word $\mathbf{w}$ of the DN test set and $m \in \{ \text{context}, \text{form} \}$. We then define a set of \emph{rank difference buckets} \[
B = \{ \pm 10^i \mid i \in \{1, \ldots, 4 \} \} \cup \{ 0 \}
\] and assign each word $\mathbf{w}$ to its closest bucket,
\[
b_{\mathbf{w},m} = \arg\min_{b \in B} | b - d_m(\mathbf{w}) |.
\]
The number of words in each so-obtained bucket can be seen
for both surface-form and context embeddings in
Figure~\ref{ranks-intra}.
To get an understanding of how different
combination functions influence the resulting embeddings,
rank differences are shown for both the single-parameter
and  gated configurations of the form-context model.

\begin{table}[t!]
  \centering
  {
\begin{tabularx}{\linewidth}{X}
\toprule
\textbf{Words with high form weight} ($\alpha \leq 0.1$) \\
 cookstown, feltham, sydenham, wymondham, cleveland, banbury, highbury, shaftesbury \\
 \midrule
\textbf{Words with high context weight} ($\alpha > 0.3$) \\
poverty, hue, slang, flax, rca, bahia, atari, snooker, icq, bronze, esso \\
\bottomrule
\end{tabularx}}
\caption{Selection of words from the DN development set where the weight of the surface-form embedding (top) or context embedding (bottom) is especially high
\label{results-word-selection}}
\end{table}

As can be seen in Figure~\ref{ranks-intra} (top), the combined architecture dramatically improves representations for approximately one third of the test words, compared to the purely surface-form-based model. These are almost exclusively words which can not or only with great difficulty be derived morphologically from any known words, including many abbreviations such as ``BMX'' and ``DDT'', but also regular words such as ``whey'', ``bled'', and ``wisdom''. While a more sophisticated model might actually be able to morphologically analyze the latter two words, our simple $n$-gram based model fails to do so. For most other words, adding context information to the surface-form model only moderately affects the quality of the obtained representations. 

As the context model assigns to most words representations
that at least
broadly capture their semantics,
only very few of its embeddings improve as much as for the
surface-form model when adding surface-form information
(Figure~\ref{ranks-inter}, bottom). However, it can be seen that
many embeddings can at least slightly be refined through
this additional information. As one might expect, the words
that profit most are those for which the provided
definitions are hard to understand and a morphological
analysis is comparatively easy, including
``parliamentarian'', ``virtuosity'' and ``drowning''. We can
also see the positive influence of designing $\alpha$ as a
function of both embeddings, i.e., of the gated model: It
does a better job at
deciding when context-based embeddings may be improved by
adding surface-form-based information.
However, it can also be seen that the representations of several words worsen when combining the two embeddings. In accordance with the observations made for the CRW dataset, this indicates that the model might further be improved by refining the composition function.

In order to gain further insight into the model's strengths and weaknesses, we finally evalute it on several subgroups of the DN test set. To this end, we categorize all nouns contained therein as either proper nouns or common nouns, further subdividing the latter category into nouns whose lemma also occurs in other frequent words (e.g. ``printing'' and ``computation'') and other nouns (e.g. ``honey'' and ``april''). Table~\ref{results-nouns} shows the performance of the form-context model for each of these word groups. Naturally, the surface-form model performs far better for words with known lemmata than for other words; it struggles the most with proper nouns as the meaning of many such nouns can not easily be derived from their surface form. Accordingly, proper nouns are the only category for which the purely context-based model performs better than the surface-form model. It is interesting to note that the improvements from combining the two embeddings using the gated model are consistent accross all categories. The largest difference between the single-parameter and the gated model can be observed for nouns whose lemma does not occur in other frequent words. This further indicates that the gated model is able to detect words which can not easily be reduced to known lemmata and, accordingly, downweights the surface-form embedding for those words.

\begin{table}
  \centering
  {
\begin{tabularx}{\linewidth}{Xrrr}
\toprule
\textbf{Model} & \textbf{Proper nouns} & \multicolumn{2}{c}{\textbf{Common nouns}} \\
& (126) & lem (79) & oth (86) \\
\midrule
surface-form & 0.03 & 0.29 & 0.12 \\
context & 0.06 & 0.09 & 0.05 \\
single-parameter & 0.10 & 0.32 & 0.11 \\
gated & 0.11 & 0.32 & 0.15 \\
\bottomrule
\end{tabularx}}
\caption{MRR of the embeddings inferred by the form-context-model and its components for proper nouns and common nouns from the DN test set. Common nouns are divided into nouns with known lemmata (lem) and those without (oth). The number of words in each group is shown in parantheses.
\label{results-nouns}}
\end{table}

\section{Conclusion and Future Work}
We have presented a  model that is capable of inferring high-quality representations for novel words by processing both the word's internal structure and words in its context. This is done by intelligently combining an embedding based on $n$-grams with an embedding obtained from averaging over all context words. Our algorithm can be trained from and combined with any preexisting word embedding model. On both the Definitional Nonce dataset and the Contextual Rare Words dataset, our model outperforms all previous approaches to learning embeddings of rare words by a large margin, even beating the embedding algorithm it was trained from on the latter dataset. Careful analysis of our combined model showed that in many cases, it is able to effectively balance out the influences of both embeddings it is composed of, allowing it to greatly improve upon representations that are either purely surface-form-based or purely context-based.
By providing a development set that complements the CRW dataset, we hope to further spur research in the area of ``few-shot learning'' for word embeddings.


While we showed that a context-dependent combination of surface-form and
context embeddings  substantially improves the
model's performance on the Definitional Nonce task, results
on the Contextual
Rare Words dataset indicate that there is still room for
further enhancement. This could potentially be achieved by
incorporating the number and informativeness of the
available contexts into the composition function; i.e., the
gate would not only be conditioned on the embeddings, but on
richer information about the context sentences. It would
also be interesting to investigate whether our model profits
from using more complex ways than averaging to obtain
surface-form and context embeddings, respectively. For example, one might introduce weights for \mbox{$n$-grams} and words depending on their contexts (i.e. the \mbox{$n$-grams} or words surrounding them). 
For scenarios in which not just one, but multiple contexts are available to infer a word's embedding, a promising extension of our model is to weight the influence of each context based on its ``definitional quality''; a similar modification was also proposed by \namecite{herbelot2017high} for their Nonce2Vec model.
Yet another interesting approach would be to integrate relative position information into our model. This could be done similar to \namecite{shaw2018selfattention} by additionally learning position embeddings and weighting the influence of context words based on those embeddings. 

\section*{Acknowledgments}
This work was funded by the European Research Council (ERC \#740516).
We would like to thank the anonymous reviewers
for their helpful comments.

\bibliography{schick}

\begin{thebibliography}{}

\bibitem[\protect\citeauthoryear{Ataman and
  Federico}{2018}]{ataman2018compositional}
Ataman, D., and Federico, M.
\newblock 2018.
\newblock Compositional representation of morphologically-rich input for neural
  machine translation.
\newblock In {\em Proceedings of the 56th Annual Meeting of the Association for
  Computational Linguistics (Volume 2: Short Papers)},  305--311.
\newblock Association for Computational Linguistics.

\bibitem[\protect\citeauthoryear{Bojanowski \bgroup et al\mbox.\egroup
  }{2017}]{bojanowski2016enriching}
Bojanowski, P.; Grave, E.; Joulin, A.; and Mikolov, T.
\newblock 2017.
\newblock Enriching word vectors with subword information.
\newblock {\em Transactions of the Association for Computational Linguistics}
  5:135--146.

\bibitem[\protect\citeauthoryear{Conneau and Kiela}{2018}]{conneau2018senteval}
Conneau, A., and Kiela, D.
\newblock 2018.
\newblock Senteval: An evaluation toolkit for universal sentence
  representations.
\newblock {\em arXiv preprint arXiv:1803.05449}.

\bibitem[\protect\citeauthoryear{Cotterell, Sch{\"u}tze, and
  Eisner}{2016}]{cotterell2016morphological}
Cotterell, R.; Sch{\"u}tze, H.; and Eisner, J.
\newblock 2016.
\newblock Morphological smoothing and extrapolation of word embeddings.
\newblock In {\em Proceedings of the 54th Annual Meeting of the Association for
  Computational Linguistics (Volume 1: Long Papers)},  1651--1660.
\newblock Association for Computational Linguistics.

\bibitem[\protect\citeauthoryear{Gerz \bgroup et al\mbox.\egroup
  }{2018}]{gerz2018language}
Gerz, D.; Vulic, I.; Ponti, E.~M.; Naradowsky, J.; Reichart, R.; and Korhonen,
  A.
\newblock 2018.
\newblock Language modeling for morphologically rich languages: Character-aware
  modeling for word-level prediction.
\newblock {\em {TACL}} 6:451--465.

\bibitem[\protect\citeauthoryear{Glorot and
  Bengio}{2010}]{glorot2010understanding}
Glorot, X., and Bengio, Y.
\newblock 2010.
\newblock Understanding the difficulty of training deep feedforward neural
  networks.
\newblock In {\em Proceedings of the Thirteenth International Conference on
  Artificial Intelligence and Statistics}, Proceedings of Machine Learning
  Research,  249--256.
\newblock PMLR.

\bibitem[\protect\citeauthoryear{Goldberg}{2016}]{goldberg2016primer}
Goldberg, Y.
\newblock 2016.
\newblock A primer on neural network models for natural language processing.
\newblock {\em Journal of Artificial Intelligence Research} 57(1):345--420.

\bibitem[\protect\citeauthoryear{Herbelot and Baroni}{2017}]{herbelot2017high}
Herbelot, A., and Baroni, M.
\newblock 2017.
\newblock High-risk learning: acquiring new word vectors from tiny data.
\newblock In {\em Proceedings of the 2017 Conference on Empirical Methods in
  Natural Language Processing},  304--309.
\newblock Association for Computational Linguistics.

\bibitem[\protect\citeauthoryear{Hochreiter and
  Schmidhuber}{1997}]{hochreiter1997long}
Hochreiter, S., and Schmidhuber, J.
\newblock 1997.
\newblock Long short-term memory.
\newblock {\em Neural Computation} 9(8):1735--1780.

\bibitem[\protect\citeauthoryear{Khodak \bgroup et al\mbox.\egroup
  }{2018}]{khodak2018carte}
Khodak, M.; Saunshi, N.; Liang, Y.; Ma, T.; Stewart, B.; and Arora, S.
\newblock 2018.
\newblock A la carte embedding: Cheap but effective induction of semantic
  feature vectors.
\newblock In {\em Proceedings of the 56th Annual Meeting of the Association for
  Computational Linguistics (Volume 1: Long Papers)},  12--22.
\newblock Association for Computational Linguistics.

\bibitem[\protect\citeauthoryear{Kingma and Ba}{2015}]{kingma2014adam}
Kingma, D., and Ba, J.
\newblock 2015.
\newblock Adam: A method for stochastic optimization.
\newblock {\em The International Conference on Learning Representations
  (ICLR)}.

\bibitem[\protect\citeauthoryear{Lazaridou \bgroup et al\mbox.\egroup
  }{2013}]{lazaridou2013compositional}
Lazaridou, A.; Marelli, M.; Zamparelli, R.; and Baroni, M.
\newblock 2013.
\newblock Compositional-ly derived representations of morphologically complex
  words in distributional semantics.
\newblock In {\em Proceedings of the 51st Annual Meeting of the Association for
  Computational Linguistics (Volume 1: Long Papers)},  1517--1526.
\newblock Association for Computational Linguistics.

\bibitem[\protect\citeauthoryear{Lazaridou, Marelli, and
  Baroni}{2017}]{lazaridou2017multimodal}
Lazaridou, A.; Marelli, M.; and Baroni, M.
\newblock 2017.
\newblock Multimodal word meaning induction from minimal exposure to natural
  text.
\newblock {\em Cognitive Science} 41:677--705.

\bibitem[\protect\citeauthoryear{Luong, Socher, and
  Manning}{2013}]{luong2013better}
Luong, T.; Socher, R.; and Manning, C.
\newblock 2013.
\newblock Better word representations with recursive neural networks for
  morphology.
\newblock In {\em Proceedings of the Seventeenth Conference on Computational
  Natural Language Learning},  104--113.

\bibitem[\protect\citeauthoryear{Mikolov \bgroup et al\mbox.\egroup
  }{2013}]{Mikolov2013}
Mikolov, T.; Chen, K.; Corrado, G.; and Dean, J.
\newblock 2013.
\newblock Efficient estimation of word representations in vector space.
\newblock {\em CoRR} abs/1301.3781.

\bibitem[\protect\citeauthoryear{Pinter, Guthrie, and
  Eisenstein}{2017}]{pinter2017mimicking}
Pinter, Y.; Guthrie, R.; and Eisenstein, J.
\newblock 2017.
\newblock Mimicking word embeddings using subword {RNNs}.
\newblock In {\em Proceedings of the 2017 Conference on Empirical Methods in
  Natural Language Processing},  102--112.
\newblock Association for Computational Linguistics.

\bibitem[\protect\citeauthoryear{Salle and
  Villavicencio}{2018}]{salle2018incorporating}
Salle, A., and Villavicencio, A.
\newblock 2018.
\newblock Incorporating subword information into matrix factorization word
  embeddings.
\newblock In {\em Proceedings of the Second Workshop on Subword/Character LEvel
  Models},  66--71.
\newblock Association for Computational Linguistics.

\bibitem[\protect\citeauthoryear{Salle, Idiart, and
  Villavicencio}{2016}]{salle2016matrix}
Salle, A.; Idiart, M.; and Villavicencio, A.
\newblock 2016.
\newblock Matrix factorization using window sampling and negative sampling for
  improved word representations.
\newblock In {\em Proceedings of the 54th Annual Meeting of the Assocation for
  Computational Linguistics (Volume 2: Short Papers)}.
\newblock Association for Computational Linguistics.

\bibitem[\protect\citeauthoryear{Shaoul and
  Westbury}{2010}]{shaoul2010westbury}
Shaoul, C., and Westbury, C.
\newblock 2010.
\newblock The westbury lab wikipedia corpus.

\bibitem[\protect\citeauthoryear{Shaw, Uszkoreit, and
  Vaswani}{2018}]{shaw2018selfattention}
Shaw, P.; Uszkoreit, J.; and Vaswani, A.
\newblock 2018.
\newblock Self-attention with relative position representations.
\newblock In {\em Proceedings of the 2018 Conference of the North American
  Chapter of the Association for Computational Linguistics: Human Language
  Technologies, Volume 2 (Short Papers)},  464--468.
\newblock Association for Computational Linguistics.

\bibitem[\protect\citeauthoryear{Wieting \bgroup et al\mbox.\egroup
  }{2016}]{wieting2016charagram}
Wieting, J.; Bansal, M.; Gimpel, K.; and Livescu, K.
\newblock 2016.
\newblock Charagram: Embedding words and sentences via character n-grams.
\newblock {\em CoRR} abs/1607.02789.

\end{thebibliography}
\bibliographystyle{aaai}
\end{document}